\documentclass{article}
\usepackage{graphicx}  % For including images
\usepackage{amsmath}   % For mathematical equations
\usepackage{hyperref}  % For hyperlinks
\usepackage{geometry}  % For adjusting page layout
\usepackage{natbib}
\setcitestyle{authoryear,open={(},close={)}} %Citation-related commands

\newcommand{\ctslogo}{\raisebox{3.4pt}{\includegraphics[scale=0.0105]{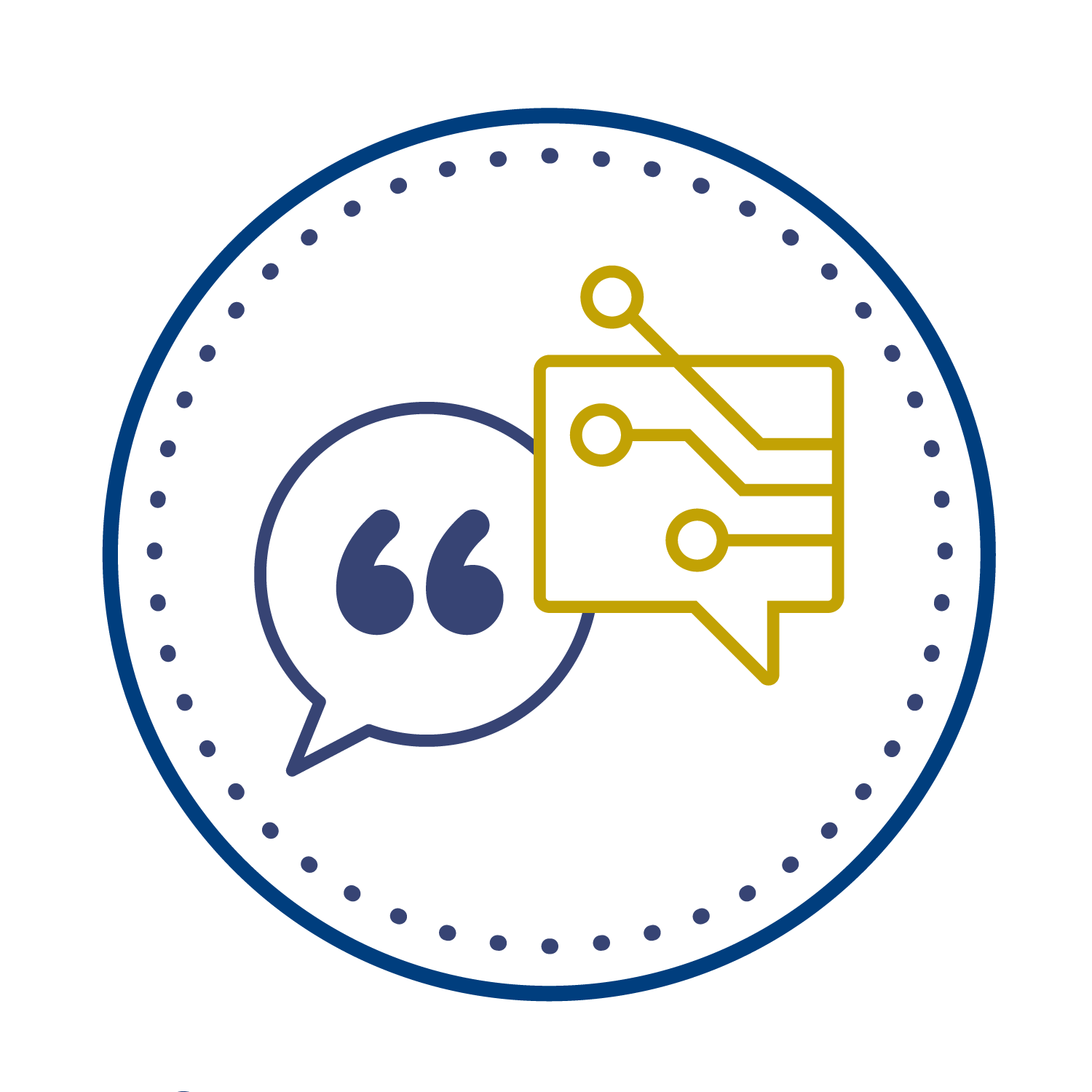}}}
\newcommand{\PAIlogo}{\raisebox{3.4pt}{\includegraphics[scale=0.080]{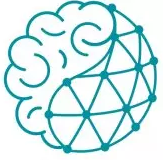}}}

\title{Edit Distances and Their Applications to Downstream Tasks in Research and Commercial Contexts}
\author{Félix do Carmo\ctslogo and Diptesh Kanojia\PAIlogo\\
        \ctslogo Centre for Translation Studies and \PAIlogo Institute for People-Centred AI,\\
        University of Surrey, United Kingdom.\\
        \texttt{\{f.docarmo, d.kanojia\}@surrey.ac.uk}
        }
\date{(Tutorial @ $16$th AMTA Conference, 2024)}

\begin{document}

\maketitle

\begin{abstract}
The tutorial describes the concept of edit distances applied to research and commercial contexts. We use Translation Edit Rate (TER), Levenshtein, Damerau-Levenshtein, Longest Common Subsequence and $n$-gram distances to demonstrate the \textit{frailty} of statistical metrics when comparing text sequences. Our discussion disassembles them into their essential components. We discuss the centrality of four editing actions: insert, delete, replace and move words, and show their implementations in openly available packages and toolkits. The application of edit distances in downstream tasks often assumes that these accurately represent work done by post-editors and real errors that need to be corrected in MT output. We discuss how imperfect edit distances are in capturing the details of this \textit{error correction} work and the implications for researchers and for commercial applications, of these uses of edit distances. In terms of commercial applications, we discuss their integration in computer-assisted translation tools and how the perception of the connection between edit distances and post-editor effort affects the definition of translator rates.
\end{abstract}

\section{Introduction}
Edit distances are a class of metrics used to quantify the similarity between two text sequences by calculating the minimum number of operations required to transform one sequence into another. These operations typically include insertion, deletion, substitution, and movement of characters or words. The application of edit distances extends beyond simple string comparison and is used extensively in evaluating machine-translated text against human references, quality estimation, and post-editing tasks.

This tutorial is targeted at researchers of machine translation and of human translation, as well as corporate members of AMTA. It focuses on the uses of edit distances, such as TER - Translation Edit Rate~\citep{Snover2006AStudyOT}, as proxies of translation effort and as informants of other downstream tasks, such as MT evaluation and post-editing, error annotation with MQM~\citep{Burchardt2013MultidimensionalQM}, quality estimation - QE~\citep{Specia2022QualityEF} and automatic post-editing - APE~\citep{doCarmo2021AReviewOT}. The application of edit distances in downstream tasks often assumes that these accurately represent work done by post-editors and real errors that need to be corrected in MT output. We will discuss how imperfect edit distances are in capturing the details of this error correction work and the implications for researchers and for commercial applications of these uses of edit distances. In terms of commercial applications, we will discuss their integration in computer-assisted translation tools and how the perception of the connection between edit distances and post-editor effort affects the definition of translator rates.

\section{Tutorial Overview}
The tutorial is divided into four parts as described below.

\paragraph{Introduction to Edit Distances and Implementations (30 mins)}
Edit distances play a crucial role in translation workflows, particularly in evaluating MT output and estimating the effort required for post-editing. In part one, we will elaborate on the statistical nature of edit distances and their different implementations and applications. We conclude with some time for questions.

\paragraph{Analyzing a Sequence of Edits (35 mins)}
In the second part, we will present a practical exercise in which a set of sentences, which simulates a sequence of edit steps with increasing complexity, is analyzed with TER and other edit distances. We will compare the details provided by TER with the actual actions performed, to assess the accuracy of this edit distance. We introduce HER~\citep{doCarmo2021EditingAA}, and discuss its utility for measuring human effort in contrast with TER. \textit{We will then allow for discussion time and for a short break of 20 mins}

\paragraph{Building a Computational Perspective (40 mins)}
The third part will be based on an exercise prepared with Python packages. We utilize a package which includes edit distances and similarity metrics (strsimpy\footnote{\url{https://pypi.org/project/strsimpy/}}), PyTER\footnote{\url{https://pypi.org/project/pyter3/}} for an implementation of TER, and SacreBLEU\footnote{\url{https://github.com/mjpost/sacrebleu}} for BLEU~\citep{papineni2002bleu}, chrF~\citep{popovic2015chrf}, and an alternative implementation of TER, for discussing statistical metrics and their sensitivity. Participants can opt to reproduce the sequence of the exercise in a Python notebook that will be shared, or to watch the step-by-step demonstration.

\paragraph{Implications for Research and Commercial Contexts (30 mins)}
Other downstream NLP tasks also use edit distances. Quality Estimation (QE) predicts the quality of MT output without requiring a reference translation. It can be used to inform decisions about whether an MT output requires post-editing and to what extent. Automatic Post-Editing (APE) systems aim to correct MT errors using computational models. However, both QE and APE face challenges related to \textit{overcorrection} and the reliance on imperfect metrics like TER. We also discuss recent collaborative efforts to merge \textit{evaluation} and \textit{correction} post-MT. The tutorial will conclude with a discussion about the implications for research and commercial applications of edit distances in cases such as use of MQM in QE and APE.

\section{Conclusion}
Edit distances provide a useful framework for evaluating and improving MT output, but they are not without limitations. Researchers and practitioners must be mindful of these challenges and continue to refine these metrics to better capture the complexities of human language. This tutorial provides a foundation for understanding these issues and offers practical tools and code resources for further exploration.

\section*{Repository and Resources}
The code and resources used in this tutorial are available on GitHub: \url{https://github.com/surrey-nlp/AMTA-EditDistances-tutorial}. The repository contains:
\begin{itemize}
    \item Python notebook demonstrating the calculation of various edit distances and metrics.
    \item Example dataset used in the tutorial for hands-on exercises.
    \item Tutorial Slides.
\end{itemize}

\section*{Speakers}
\textbf{Dr Félix do Carmo} is a Senior Lecturer at the Centre for Translation Studies of the University of Surrey, where he teaches and researches the application of technologies to translation work processes, with a focus on their ethical and professional implications. He is a Fellow of the Surrey Institute for People-Centred Artificial Intelligence, and an Expert member of the Surrey Future of Work Research Centre. He worked for more than 20 years in Porto, Portugal, as a translator, translation company owner and university lecturer, and he was awarded a post-doctoral research fellowship to work at Dublin City University.\\
Google Scholar: https://scholar.google.com/citations?user=MF9s2xsAAAAJ

\textbf{Dr Diptesh Kanojia} is a Lecturer at Institute for People-centred Artificial Intelligence at the University of Surrey where he leads the taught module on Natural Language Processing. He works on various Natural Language Processing sub-domains and his current research interests are quality estimation (QE), automatic post-editing (APE), social NLP and low-resource NLP scenarios. He leads an industry project sponsored by eBay Inc. and two projects sponsored by the European Association for Machine Translation. He also co-leads an NHS sponsored project on clinical question answering for orthopaedic surgery patients, and co-organises the Conference for Machine Translation (WMT) Shared tasks on QE and APE. He is passionate about teaching, diversity \& inclusion, and applying research to real-world applications.\\
Google Scholar: https://scholar.google.com/citations?user=UNCgCAEAAAAJ  

\section*{Acknowledgments}
We would like to thank the organizers of the 16th Biennial Conference of the Association for Machine Translation in the Americas for the opportunity to present this tutorial.

\bibliographystyle{abbrvnat}
\bibliography{references}

\begin{thebibliography}{7}
\providecommand{\natexlab}[1]{#1}
\providecommand{\url}[1]{\texttt{#1}}
\expandafter\ifx\csname urlstyle\endcsname\relax
  \providecommand{\doi}[1]{doi: #1}\else
  \providecommand{\doi}{doi: \begingroup \urlstyle{rm}\Url}\fi

\bibitem[Burchardt(2013)]{Burchardt2013MultidimensionalQM}
A.~Burchardt.
\newblock Multidimensional quality metrics: a flexible system for assessing translation quality.
\newblock In \emph{Proceedings of Translating and the Computer 35}, 2013.

\bibitem[do~Carmo(2021)]{doCarmo2021EditingAA}
F.~do~Carmo.
\newblock Editing actions: a missing link between translation process research and machine translation research.
\newblock In \emph{Explorations in empirical translation process research}, pages 3--38. Springer, 2021.

\bibitem[do~Carmo et~al.(2021)do~Carmo, Shterionov, Moorkens, Wagner, Hossari, Paquin, Schmidtke, Groves, and Way]{doCarmo2021AReviewOT}
F.~do~Carmo, D.~Shterionov, J.~Moorkens, J.~Wagner, M.~Hossari, {\'E}.~Paquin, D.~Schmidtke, D.~Groves, and A.~Way.
\newblock A review of the state-of-the-art in automatic post-editing.
\newblock \emph{Machine Translation}, 35:\penalty0 101--143, 2021.

\bibitem[Papineni et~al.(2002)Papineni, Roukos, Ward, and Zhu]{papineni2002bleu}
K.~Papineni, S.~Roukos, T.~Ward, and W.-J. Zhu.
\newblock Bleu: a method for automatic evaluation of machine translation.
\newblock In \emph{Proceedings of the 40th annual meeting of the Association for Computational Linguistics}, pages 311--318, 2002.

\bibitem[Popovi{\'c}(2015)]{popovic2015chrf}
M.~Popovi{\'c}.
\newblock chrf: character n-gram f-score for automatic mt evaluation.
\newblock In \emph{Proceedings of the tenth workshop on statistical machine translation}, pages 392--395, 2015.

\bibitem[Snover et~al.(2006)Snover, Dorr, Schwartz, Micciulla, and Makhoul]{Snover2006AStudyOT}
M.~Snover, B.~Dorr, R.~Schwartz, L.~Micciulla, and J.~Makhoul.
\newblock A study of translation edit rate with targeted human annotation.
\newblock In \emph{Proceedings of the 7th Conference of the Association for Machine Translation in the Americas: Technical Papers}, pages 223--231, 2006.

\bibitem[Specia et~al.(2022)Specia, Scarton, and Paetzold]{Specia2022QualityEF}
L.~Specia, C.~Scarton, and G.~H. Paetzold.
\newblock \emph{Quality estimation for machine translation}.
\newblock Springer Nature, 2022.

\end{thebibliography}

\end{document}